\documentclass{article}

\PassOptionsToPackage{numbers, compress}{natbib}
\usepackage[final]{nips_2016}

\usepackage[utf8]{inputenc} 
\usepackage[T1]{fontenc}    
\usepackage{url}            
\usepackage{booktabs}       
\usepackage{amsfonts}       
\usepackage{nicefrac}       
\usepackage{microtype}      
\usepackage{amssymb}
\usepackage{epsfig}
\usepackage{amsfonts}
\usepackage{mathrsfs}
\usepackage{amsmath}
\usepackage{graphicx}
\usepackage{color}
\usepackage{caption}
\usepackage{picinpar}
\usepackage{booktabs}
\usepackage{float}
\usepackage{url}
\usepackage{dsfont}
\usepackage{multirow}
\usepackage{subfloat}
\usepackage{slashbox}
\usepackage{array}
\usepackage{subfigure}
\usepackage[none]{hyphenat}
\usepackage[ruled,vlined,linesnumbered]{algorithm2e}
\usepackage[english]{babel}
\newcommand{\tabincell}[2]{\begin{tabular}{@{}#1@{}}#2\end{tabular}}
\def\BibTeX{{\rm B\kern-.05em{\sc i\kern-.025em b}\kern-.08em
    T\kern-.1667em\lower.7ex\hbox{E}\kern-.125emX}}
\usepackage{commath}
\usepackage{color}
\usepackage{lineno}
\usepackage[flushleft]{threeparttable}
\usepackage[utf8]{inputenc}
\usepackage[english]{babel}
\usepackage{amsthm}
\usepackage{bm}

\usepackage{hyperref}
\title{Deep Stochastic Configuration Networks with Universal Approximation Property}

\author{
  Dianhui~Wang\thanks{Corresponding author.}\:\:\:\:\:\:\:\:\:\:\:\:\:\:\:\:\:\:Ming Li\\
  Department of Computer Science and Information Technology\\
  La Trobe University,
  Melbourne, VIC 3086, Australia \\
  \texttt{Email:dh.wang@latrobe.edu.au} \\
}

\begin{document}

\maketitle

\begin{abstract}
This paper develops a randomized approach  for incrementally building deep neural networks, where a supervisory mechanism is proposed to constrain the random assignment of the  weights and biases, and all the hidden layers have direct links to the output layer. A fundamental result on the universal approximation property is established for such a class of randomized leaner models, namely deep stochastic configuration networks (DeepSCNs). A learning algorithm is  presented to implement DeepSCNs with either specific architecture or self-organization. The read-out weights attached with all direct links from each hidden layer to the output layer are evaluated by the least squares method.  Given a set of training examples, DeepSCNs can  speedily produce a learning representation, that is,  a collection of random basis functions with the cascaded inputs together with the read-out weights. An empirical demonstration
on a function approximation and a case study are carried out to demonstrate some properties of the proposed deep learner model.
\end{abstract}
\section{Introduction}
In the past decade, deep neural networks (DNNs) have received considerable attention due to its great potential for dealing with computer vision, pattern recognition, natural language processing, etc. \cite{lecun, schmi, goodf}. The success of deep learning is attributed to its  representation capability for visual data \cite{ hinto,bengi}. In view of some empirical evidence, DNNs are becoming increasingly popular because of a hypothesis that a deep learner model can be  more effective and efficient in data representation aspect, compared to  shallow models. It seems that DNNs have  good potential in learning higher-level abstractions (or multiple levels of features), which sounds difficult for models with shallow architecture. A formal analysis on  the representation power and learning complexity of DNNs can be found in \cite{eldan}. In short, the community has reached a common sense that DNNs are much more expressive than the shallow ones.

DNNs became so popular due to their learning representation power for complex data, however, there are some issues in the design and implementation. For instance, how to smartly choose the architecture of the DNN, improve the learning speed and reduce the computational burden. Usually, one configures a DNN manually based on the trial-and-error performance without technical guidance, which may be time-consuming and potentially impacts the effectiveness of the resulting model. Indeed,  DNNs with moderate depth but fewer number of hidden nodes at each layer cannot exhibit sufficient learning ability, while a large sized architecture (i.e. with quite a few layers and hidden nodes) may cause over-fitting that degrades the generalization capability.  In \cite{alvar}, Alvarez and Salzmann introduced an approach to automatically determine the number of nodes at each layer of the DNN by using group sparsity regularizer. Their method can result in a compact width setting for each layer, but with the prerequisite that the network's depth is determined in advance. Except for the architecture setting problem, fast training algorithms for DNNs are significant as well. As can be seen that training multilayer perceptrons by using the  back-propagation algorithm suffers from several issues such as the weight initialisation, local minima and sensitivity of the learning performance with respect to the learning rate setting. Empirically, this gradient-based learning method could not produce meaningful or interpretable internal representations from each hidden outputs \cite{rumel}.  Therefore, some advanced deep learning techniques have been developed to overcome these obstacles \cite{hintonfast}. To the best of our knowledge, there are few deep learning techniques in literature addressing  the design and fast implementation simultaneously. In \cite{arora}, the authors explored the learning capability of autoencoders with random weights scoped in [-1,1], showing the value of the randomized methods in building DNNs with much less computational complexity. Recently, some theoretical results on the universal classification strategy using DNN with random weights were reported in \cite{girye}, where the authors mathematically showed that DNNs with random Gaussian weights perform a distance-preserving embedding of the data.
 Motivated by constructive approaches for building shallow neural networks \cite{barro, kwok} and randomized methods for fast learning \cite{galla,broom,pao, sutto,igeln, gorba, liwan, scard}, we make efforts to  develop randomized learner model with deep architecture under the  philosophy: `one should walk on the earth rather than in the air'!

This paper further develops the stochastic configuration networks (SCNs) proposed in  \cite{wang} and establishes the universal approximation property for its deep version, termed as DeepSCNs.  From an algorithmic perspective, a DeepSCN starts with a small sized network (e.g, one hidden layer with one hidden node), and stochastically configures its nodes for the current layer until a certain termination criterion is met, then continues to add the next  hidden layer by repeating the same procedures, and  keeps proceeding to deeper layers until an acceptable error tolerance is achieved. As the constructive process proceeds, the hidden parameters are randomly assigned under a supervisory mechanism, while the read-out weights linking the hidden nodes to the output nodes are calculated by the least squares method \cite{lanca}. Our basic idea behind the DeepSCN is to successively approximate a target function, and the random basis functions for each hidden layer are generated under a set of inequality constraints. Once starting to add a new hidden layer, the obtained random bases remain unchanged and the last error function acts as a new target function to be approximated. An immediate benefit from DeepSCNs lies in the links between all hidden nodes and the output nodes, which allow  end-users to manipulate  layer-wise approximations, in other words, one can make options to use these nodes freely  after off-line building an accurate DeepSCN model. Furthermore, DeepSCN can produce more rich learning representations, which provide an alternative solution for fast building autoencoders. In a nutshell, as results of deep learning for DeepSCNs,  a set of data dependent random basis functions are generated for  signal representation with piece-wise `resolutions'.

The remainder of this paper is organized as follows: Section 2 details our DeepSCNs framework with  theoretical fundamentals on the universal approximation property and algorithm description.  Section 3 reports some simulation results with comparisons against SCNs. Section 4 concludes this paper and provides some further research directions.

\section{Deep Stochastic Configuration Networks}
This section provides some fundamentals of DeepSCNs, including a  theoretical result on the universal approximation property, algorithm description and some technical remarks.
\subsection{Universal Approximation Property}
Let $\mathcal{L}_{2}(D)$ denote the space of all Lebesgue-measurable vector-valued functions $\mathcal{F}=[f_1,f_2,\ldots,f_m]:R^{d}\rightarrow R^{m}$ on a compact set $D\subset R^{d}$, with the $\mathcal{L}_{2}$ norm defined as
\begin{equation}\label{multiple_lp}
  \|\mathcal{F}\|:=\left(\sum_{q=1}^{m}\int_{D}|f_q(x)|^2dx\right)^{1/2}<\infty,
\end{equation}
and inner product defined as
\begin{equation}\label{multiple_lp}
  \langle \mathcal{F},\mathcal{G}\rangle:=\sum_{q=1}^{m}\langle f_q,g_q\rangle=\sum_{q=1}^{m}\int_{D}f_q(x)g_q(x)dx,
\end{equation}
where $\mathcal{G}=[g_1,g_2,\ldots,g_m]:R^{d}\rightarrow R^{m}$.

Given a target function $\mathcal{F}:R^{d}\rightarrow R^{m}$, suppose a DeepSCN with $n$ hidden layers and each layer has $L_k$ hidden nodes ($k=1,2,\ldots,n$), has been constructed, that is,
\begin{equation}\label{random_bases}
  \mathcal{F}^{(n)}_{S_n}(x)=\sum_{k=1}^{n}\sum_{j=1}^{L_k}\beta^{(k)}_j\phi_{k,j}\Big(x^{(k-1)};w^{(k-1)}_j,b^{(k-1)}_j\Big),
\end{equation}
where $S_n=\{L_1,L_2,\ldots,L_n\}$ represents the set of the hidden node numbers, $w^{(k-1)}_j$ and $b^{(k-1)}_j$ stand for the hidden parameters within the $k$-th hidden layer, $\phi_{k,j}$ is the activation function used in the $k$-th hidden layer, $x^{(0)}=x$ and $x^{(k)}=\Phi\Big(x^{(k-1)};W^{(k-1)},B^{(k-1)}\Big)=[\phi_{k,1},\phi_{k,2},\ldots,\phi_{k,L_{k}}]$, with $W^{(k-1)}=[w^{(k-1)}_1,w^{(k-1)}_2,\ldots,w^{(k-1)}_{L_k}]$ and $B^{(k-1)}=[b^{(k-1)}_1,b^{(k-1)}_2,\ldots,b^{(k-1)}_{L_k}]$, $k=1,2,\ldots,n$.
Then, the residual error function is defined by $\mathcal{E}^{(n)}_{S_n}=\mathcal{F}-\mathcal{F}^{(n)}_{S_n}=[\mathcal{E}^{(n)}_{S_n,1},\mathcal{E}^{(n)}_{S_n,2},\ldots,\mathcal{E}^{(n)}_{S_n,m}]$.

Denoted by $  \mathcal{F}^{(n)}_{p}(x)$ as a DeepSCN model with $p$ nodes at the $n$-th layer, that is,
\begin{equation}\label{prandom_bases}
  \mathcal{F}^{(n)}_{p}(x)=\sum_{k=1}^{n}\sum_{j=1}^{p}\beta^{(k)}_j\phi_{k,j}\Big(x^{(k-1)};w^{(k-1)}_j,b^{(k-1)}_j\Big),
\end{equation}
where $p=1,2,..., L_n$. Specifically, $\mathcal{E}^{(1)}_{0}=\mathcal{F}$ and $\mathcal{E}^{(n+1)}_{0}=\mathcal{E}^{(n)}_{L_n}$, $n=1,2,\ldots.$
\vspace{1.25mm}

\textbf{Theorem 1.} Suppose that span($\Gamma$) is dense in $\mathcal{L}_2$ space and $\forall \phi\in \Gamma$, $0<\|\phi\|<c$ for some $c\in R^{+}$. Given $0<r<1$ and a nonnegative decreasing sequence $\{\mu_l\}$ with $\lim_{l\rightarrow+\infty}\mu_l=0$ and $\mu_l< (1-r)$. For $n=1,2\ldots$, and $j=1,2,\ldots,L_n$, denoted by
\begin{equation}
\delta_{j,q}^{(n)}=(1-r-\mu_{j})\|\mathcal{E}^{(n)}_{j-1,q}\|^2, q=1,2,\ldots,m.
\end{equation}
Stochastically configuring the $j$-th hidden node $\phi_{n,j}$ within the $n$-th hidden layer ($j=1,2,\ldots,L_n$) to satisfy the following inequalities:
\begin{equation}\label{step2}
\langle \mathcal{E}^{(n)}_{j,q},\phi_{n,j}\rangle^2\geq c^2\delta_{j,q}^{(n)}, q=1,2,\ldots,m.
\end{equation}
Fix the random basis functions $\phi_{n,1},\phi_{n,2},\ldots,\phi_{n,L_n}$ and start to add the first hidden node $\phi_{n+1,1}$ in the $(n+1)$-th hidden layer according to the following inequalities
\begin{equation}\label{step3}
\langle \mathcal{E}^{(n)}_{L_n,q},\phi_{n+1,1}\rangle^2\geq c^2\delta_{L_n,q}^{(n)},q=1,2,\ldots,m.
\end{equation}
Keep adding new hidden nodes within the $(n+1)$-th hidden layer based on (\ref{step2}), followed by generating the first hidden node in a new hidden layer via (\ref{step3}). After adding one hidden node (either in the present hidden layer or starting a new hidden layer), the read-out weights are evaluated by the least squares method, that is,
\begin{equation}\label{step4}
\beta^{*}=\arg \min_{\beta}\|\mathcal{F}-\sum_{k=1}^{n}\sum_{j=1}^{L_k}\beta^{(k)}_j\phi_{k,j}\|^2.
\end{equation}
Then, we have $\lim_{n\rightarrow +\infty}\|\mathcal{F}-\mathcal{F}^{(n)}_{S_n}\|=0$, where $\mathcal{F}^{(n)}_{S_n}$ is defined by (\ref{random_bases}).\\
\vspace{1.25mm}
\textbf{Proof.} It is easy to obtain that
\begin{equation}\label{step10}
\|\mathcal{E}^{(n)}_{L_{n}}\|^2\leq(r+\mu_{L_n})\|\mathcal{E}^{(n)}_{L_n-1}\|^2.
\end{equation}
The error after adding one hidden node in the $(n+1)$-hidden layer can be expressed as
\begin{eqnarray}
\|\mathcal{E}^{(n+1)}_{1}\|^2=\|\mathcal{E}^{(n)}_{L_{n}}-\beta^{(n+1)}_1\phi_{n+1,1}\|^2,
\end{eqnarray}
where $\phi_{n+1,1}$ and $\beta^{(n+1)}_1$ are obtained by using (\ref{step3}) and (\ref{step4}) with , respectively.
To identify the relationship between $\mathcal{E}^{(n+1)}_{1}$  and $\mathcal{E}^{(n)}_{L_{n}}$, we denote $\bar{\beta}^{(n+1)}_{1}=[\bar{\beta}^{(n+1)}_{1,1}, \ldots,\bar{\beta}^{(n+1)}_{1,m}]^{\mathrm{T}}$, where
\begin{equation}\label{step9}
\bar{\beta}^{(n+1)}_{1,q}=\frac{\langle \mathcal{E}^{(n)}_{L_n,q},\phi_{n+1,1}\rangle}{\|\phi_{n+1,1}\|^2}, q=1,2,\ldots,m,
\end{equation}
Replacing $\beta^{(n+1)}_1$ by $\bar{\beta}^{(n+1)}_{1}$, with $\bar{\beta}^{(n+1)}_{1,q}$ evaluated by (\ref{step9}), we have
\begin{eqnarray}\label{step11}
\|\mathcal{E}^{(n+1)}_{1}\|^2&\leq&\|\mathcal{E}^{(n)}_{L_{n}}-\bar{\beta}^{(n+1)}_{1}\phi_{n+1,1}\|^2\nonumber\\
&=&\|\mathcal{E}^{(n)}_{L_{n}}\|^2-\frac{\sum_{q=1}^{m}\langle \mathcal{E}^{(n)}_{L_{n},q},\phi_{n+1,1}\rangle^2}{\|\phi_{n+1,1}\|^2}\nonumber\\
&\leq & \|\mathcal{E}^{(n)}_{L_{n}}\|^2.
\end{eqnarray}
With the help of (\ref{step10}) and (\ref{step11}), we can obtain
\begin{eqnarray}\label{step20}
\|\mathcal{E}^{(n+1)}_{L_{n+1}}\|^2&\leq&\|\mathcal{E}^{(n+1)}_{1}\|_{2}^2\leq \|\mathcal{E}^{(n)}_{L_{n}}\|^2 \nonumber\\
&\leq&\prod_{k=2}^{L_n}(r+\mu_{k})\|\mathcal{E}^{(n)}_{1}\|^2\nonumber\\
&\leq& (r+\mu_{1})^{(\sum_{k=1}^nL_k)-1}\|\mathcal{E}^{(1)}_{1}\|^2,
\end{eqnarray}
which implies the universal approximation property because  $r+\mu_{1}<1$ and $\sum_{k=1}^nL_k$ turns to be infinity as $n$ keeps increasing. This completes the proof.

\begin{figure}[htbp!]
\centering
\includegraphics[width=0.788\textwidth]{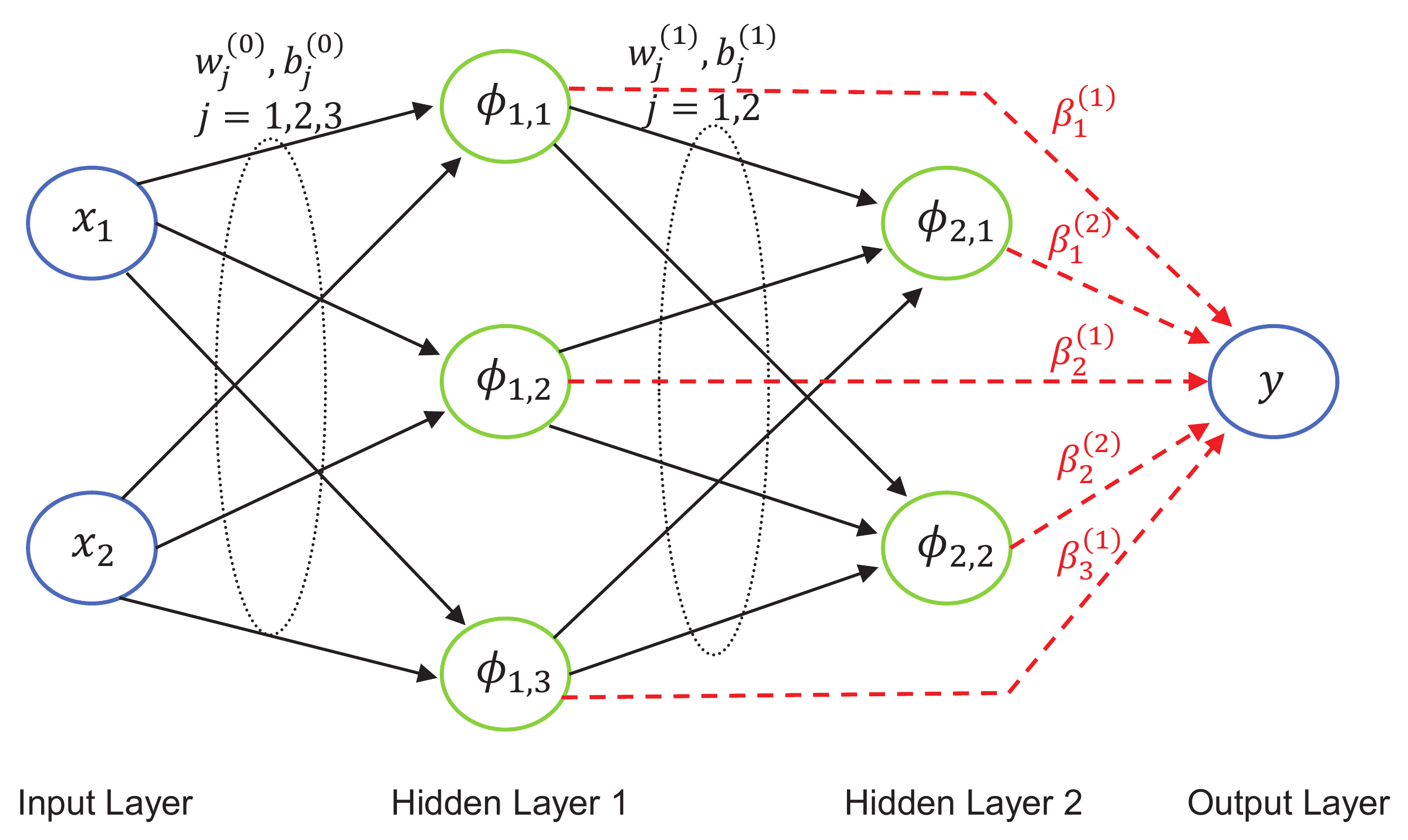}
\caption{A typical DeepSCN with the architecture: 2-3-2-1, i.e., two input nodes, two hidden layers (three hidden nodes in the first hidden layer, two hidden nods in the second hidden layer), and one output node. Read-out weights (trainable) are shown as red dashed lines, whilst hidden weights and biases (stochastically configured) are shown as fixed lines.} \label{DSCN_structure}
\end{figure}
The idea of learning internal representation can be traced back to 80s \cite{rumel}. In the past decades, many exciting researches and progresses have been reported in \cite{bengi}, where attempts are made to  meet the following three objectives: fidelity, sparsity and interpretability. Learner models can appear in different forms but essentially they must share the universal approximation property so that the fidelity can be achieved \cite{cyben, horni}. Theorem 1  ensures that DeepSCNs fit the essential prerequisite well for learning representation. Due to the nature of DeepSCNs, learning  representation becomes quite straightforward but less interpretable. The outcomes from the first hidden layer contain two parts: a set of random basis functions with the original inputs, and associated read-out weights between the first hidden layer and the output layer. Apparently, one can define a learning representation from the first hidden layer, which characterizes the main feature of the target function. Next, the outputs from the first hidden layer, as a new set of inputs, are fed into the next hidden layer to constructively generate a set of random basis functions. Again, one can obtain a refined learning representation from the second hidden layer. This procedure keeps going on until some termination criterion is satisfied. Figure 1 shows a typical DeepSCN with $n=2$, $L_1=3$, and $L_2=2$, $d=2$, $m=1$.

It should be highlighted that DeepSCNs advocate fully connections between each hidden layer and the output layer, which supports a multi-resolution of  learning representation rather than using only the compositional expression from the last hidden layer as done in the standard multilayer perceptrons.   Such a learning representation offers end-users more flexibility, for instance, some nodes at certain layer can be discarded according to some criteria, leading to a new learning representation with sparsity. After generating the random basis functions, we can employ some regularization models such as Lasso \cite{lasso} to refine the learning representation, which can be done in the layer-wise manner or at one pass.  In essence, the universal approximation property ensures the capacity of DeepSCNs for both data modelling and signal representation. Thus, DeepSCNs can be employed as either predictive models or feature extractors.

\subsection{Algorithm Description}
Given a training dataset with inputs $X=\{x_1,x_2,\ldots,x_N\}$, $x_i=[x_{i,1},\ldots,x_{i,d}]^\mathrm{T}\in R^{d}$ and outputs $T=\{t_1,t_2,\ldots,t_N\}$, where $t_i=[t_{i,1},\ldots,t_{i,m}]^\mathrm{T}\in R^{m}$, $i=1,\ldots,N$. Denote $\mathcal{E}^{(n)}_{L_n-1}=\mathcal{E}^{(n)}_{L_n-1}(X)=[\mathcal{E}^{(n)}_{L_n-1,1}(X),\ldots,\mathcal{E}^{(n)}_{L_n-1,m}(X)]^\mathrm{T}$ as the corresponding residual error vector before the $L_n$-th new hidden node of the $n$-th hidden layer is added, where $\mathcal{E}^{(n)}_{L_n-1,q}(X)=[\mathcal{E}^{(n)}_{L_n-1,q}(x_1),\ldots,\mathcal{E}^{(n)}_{L_n-1,q}(x_N)]\in R^N$, $q=1,2,\ldots,m$. \\
After adding the $L_n$-th hidden node in the $n$-th hidden layer, we get
\begin{equation}\label{hidden}
h_{L_n}^{(n)}:=h_{L_n}^{(n)}(X)=[\phi_{n,L_n}(x^{(n-1)}_1),\ldots,\phi_{n,L_n}(x^{(n-1)}_N)]^\mathrm{T},
\end{equation}
where $\phi_{n,L_n}(x^{(n-1)}_i)$ is used to simplify $\phi_{n,L_n}(x^{(n-1)}_i;w^{(n-1)}_j,b^{(n-1)}_j)$, and $x^{(0)}_i=x_i=[x_{i,1},\ldots,x_{i,d}]^\mathrm{T}$, $x^{(n-1)}_i=\Phi(x^{(n-2)};W^{(n-1)},B^{(n-1)})$ for $n\geq2$.

Let $H^{(n)}_{L_n}=[h_1^{(n)},h_2^{(n)},\ldots,h_{L_n}^{(n)}]$ represent the hidden layer output matrix and denote a temporary variable $\theta^{(n)}_{L_n,q}$ ($q=1,2,\ldots,m$) to make it convenient in the followed algorithm description (pseudo-codes).
\begin{eqnarray}\label{factor1}
\theta^{(n)}_{L_n,q}=\frac{\langle\mathcal{E}^{(n)}_{L_n-1,q},h_{L_n}^{(n)}\rangle_{l_2}^2}{\langle h_{L_n}^{(n)},h_{L_n}^{(n)}\rangle_{l_2}}-(1-r)\langle\mathcal{E}^{(n)}_{L_n-1,q}\mathcal{E}^{(n)}_{L_n-1,q}\rangle_{l_2},
\end{eqnarray}
where $\langle \cdot,\cdot\rangle_{l_2}$ denotes the dot product and we omit the argument $X$ in $\mathcal{E}^{(n)}_{L_n-1,q}$ and $h_{L_n}^{(n)}$.

\begin{algorithm}[htbp!]
\footnotesize
\SetAlgoLined
\SetKwInOut{Input}{Input}\SetKwInOut{Output}{Output}
\caption{Deep Stochastic Configuration Networks}
\label{Algorithm DSCN}
\DontPrintSemicolon
\Input{Training inputs $X=\{x_1,x_2,\ldots,x_N\}$, $x_i\in R^{d}$, outputs $T=\{t_1,t_2,\ldots,t_N\}$, $t_i\in R^{m}$; The maximum number of hidden layers, $M$; The maximum number of hidden nodes within the $n$-th hidden layer, $L_{max}^{(n)}$, $1\leq n\leq M$; The expected error tolerance $\epsilon$; The maximum times of random configuration $T_{max}$; Two sets of scalars $\Upsilon\!=\{\lambda_{1},\ldots,\lambda_{end}\}$, $\mathcal{R}=\{r_{1},\ldots,r_{end}\}$}
\Output{A DeepSCN Model}
\BlankLine
\textbf{Initialization}: $\mathcal{E}^{(1)}_0:=[t_1,t_2,\ldots,t_N]^\mathrm{T}$, $\mathcal{H},\Omega,W:=[\:\:]$;\\
\While{$n\leq M$ and $\|\mathcal{E}^{(1)}_0\|_{F}>\epsilon$}{
 \While{$L_n\leq L_{max}^{(n)}$ and $\|\mathcal{E}^{(1)}_0\|_{F}>\epsilon$}{
   \For{$\lambda \in \Upsilon$}{
     \For{$r\in \mathcal{R}$}{
       \For{$k=1,2\ldots,T_{max}$}{
        Randomly assign $\omega_{L_n}^{(n-1)}$ and $b_{L_n}^{(n-1)}$ from $[-\lambda,\lambda]^d$ and $[-\lambda,\lambda]$;\\
        Calculate $h_{L_n}^{(n)}$, $\theta^{(n)}_{L_n,q}$ by (\ref{hidden}), (\ref{factor1});\\
        \eIf{$\mbox{min}\{\theta^{(n)}_{L_n,1},\ldots,\theta^{(n)}_{L_n,m}\}\geq 0$}{
        \textbf{Save} $w_{L_n}^{(n-1)}$ and $b_{L_n}^{(n-1)}$ in $W$, $\theta^{(n)}_{L_n}=\sum_{q=1}^{m}\theta^{(n)}_{L_n,q}$ in $\Omega$;}
           {go back to \textbf{Step 5};}
           }
       \eIf {$W$ is not empty}   {Find $w_{L_n}^{(n-1)*}$, $b_{L_n}^{(n-1)*}$ maximizing $\theta^{(n)}_{L_n}$, set $H^{(n)}_{L_n}\!=\![h_{1}^{(n)*},\ldots,h_{L_n}^{(n)*}]$; \\ \textbf{Break} (go to \textbf{Step 22}); }
       {\textbf{Continue}: go to \textbf{Step 5};}
       }
   }
   Set $\mathcal{H}\!:=\![\mathcal{H},H^{(n)}_{L_n}]$, $\beta^{*}\!=\!\mathcal{H}^{\dagger}T$, $\mathcal{E}^{(n)}_{L_n}\!=\!\mathcal{H}\beta^{*}\!-\!T$, $\mathcal{E}^{(n)}=\mathcal{E}^{(n)}_{L_n}$; \\
   Renew $\mathcal{E}^{(1)}_0:=\mathcal{E}^{(n)}_{L_n}$, $L_n:=L_n+1$;
 }
 Set $\mathcal{E}^{(n+1)}_{0}=\mathcal{E}^{(n)}$, $\Omega, W=[\:\:]$;
}
\textbf{Return}: $\beta^{*}$,$\omega^{*}$,$b^{*}$
\end{algorithm}

Whilst building the DeepSCN model, we need to take the over-fitting into account. Thus, in practice,  a validation dataset is used to monitor the learning process and prevent the occurrence of the over-fitting. In this way, an appropriate architecture of the DeepSCN model can be automatically determined. Alternatively, end-users can specify the number of nodes for each layer and also the number of the layers in advance. However, this pre-defined configuration method has no guarantee to perform a good learning representation.

To speed up the implementation of DeepSCNs, the generation of the random basis functions at each hidden layer can be done in batch mode. That is, we can select a set of generative nodes that meet the constraint (\ref{step2}). Also, for data modelling tasks with high-dimensional outputs, the condition (\ref{step2}) should be replaced by satisfying $\theta^{(n)}_{L_n}=\sum_{q=1}^{m}\theta^{(n)}_{L_n,q}\ge 0$. This relaxed constraint on the random assignment of the weights and biases still ensures the universal approximation property.

For regression problems, one may be more interested in the predictability of a learner model rather than its learning capability. Unfortunately, it is almost impossible to directly establish the correlation between these two performance. That is,  a better learning performance does not always imply a sound generalization. In this regard, consistency concept becomes a meaningful metric to assess the goodness of learning machines. From our hands-on experience, DeepSCNs demonstrate favourable consistency performance. For classification problems, DeepSCNs can be used directly to put a class label for a testing data by using  linear discriminate analysis. Alternatively, we can use DeepSCNs as feature extractors to generate a collection of samples in the DeepSCN feature space, followed by using variety of feature-classifiers to implement classification.

\section{Empirical Demonstration}
This section reports some simulation results to illustrate advantages of DeepSCNs compared to  SCNs in terms of the approximation accuracy and algebraic property (rank deficiency) of the joint hidden output matrix. Here, we also provide a robustness analysis on DSCN algorithm with respect to the learning parameter setting. The following real-valued function defined over [0,1] is employed in the simulation study:
\begin{eqnarray*}
f(x) \!\!\!\!&=& \!\!\!\!0.2e^{-(10x-4)^{2}}\!\!+0.5e^{-(80x-40)^{2}}\!\!+0.3e^{-(80x-20)^{2}}.
\end{eqnarray*}
In this study, we take 1000 points (randomly generated) as the training data and another 1000 points (randomly generated) as the test data in our experiments. The sigmoidal activation function $g(x)=1/(1+\exp(-x))$ is used for all hidden nodes.

\subsection{Approximation Capability}
Figure 2 depicts  the training and test performance, respectively, where  $L=200$ for SCN,  $M=4$ and  $L_{max}^{(n)}=50$ for each layer in DeepSCN.   It is clear that DeepSCN  approximates the target function within a given tolerance more faster than SCN. Also, the consistency relationship between learning and generalization can be observed. As a matter of fact, both shallow and deep SCNs share this nice consistency property. It should be clearly pointed out that our statement on the consistency comes from a large number of experimental observations, and there is no theoretical justification available so far.   Figure 2 shows three independent trials, which indicate  that DeepSCN outperforms SCN in terms of effectiveness. In this test, we set $r_k=1-10^{-k},k=1,2,...,7$ in Step 16 of DSCN algorithm. With different settings, the performance on both effectiveness and consistency could be further verified by the results from Figure 4.
\begin{figure}[h!]
\centering
\subfigure[Training]{\includegraphics[width=0.46\textwidth]{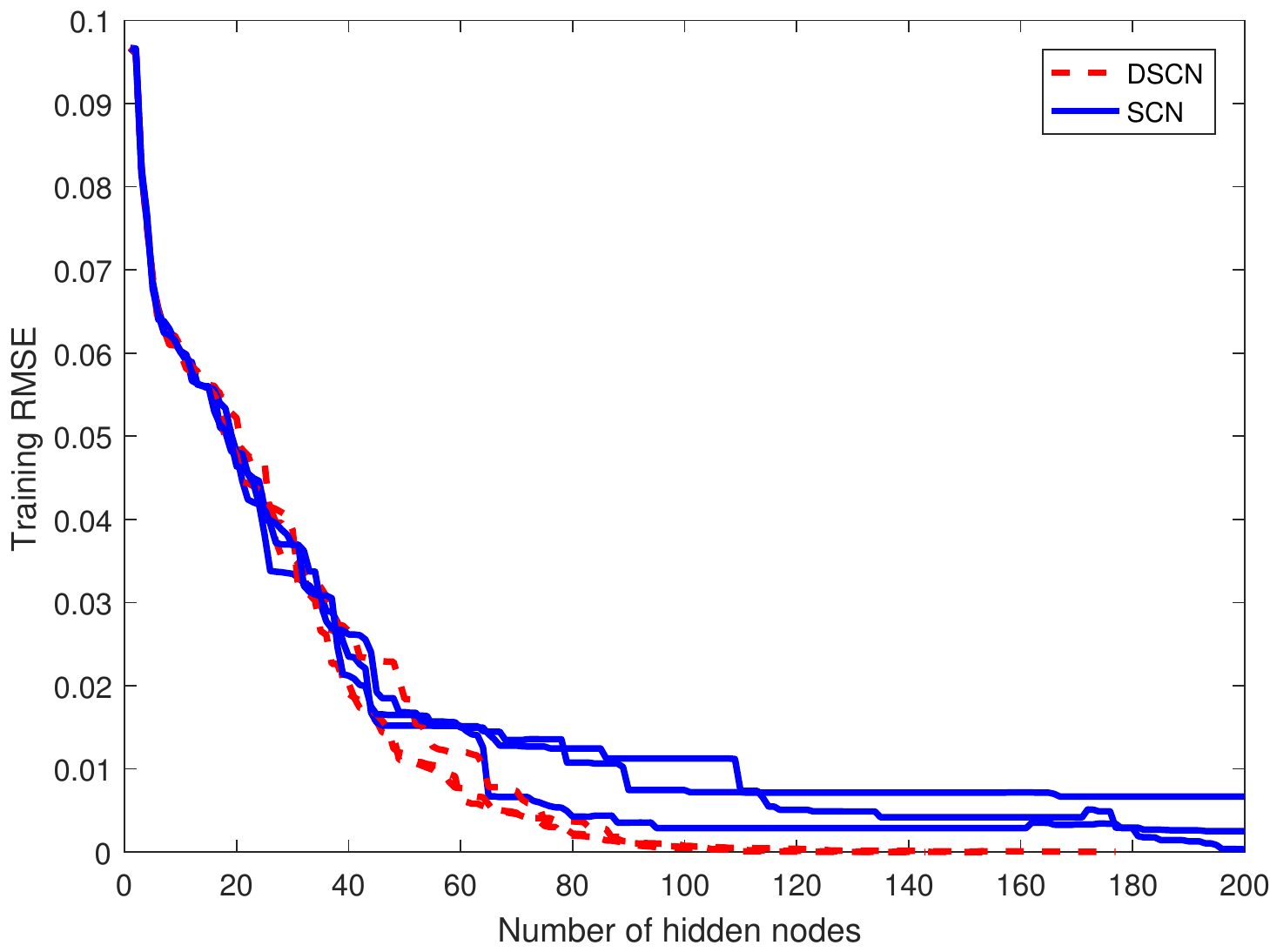}}
\subfigure[Test]{\includegraphics[width=0.46\textwidth]{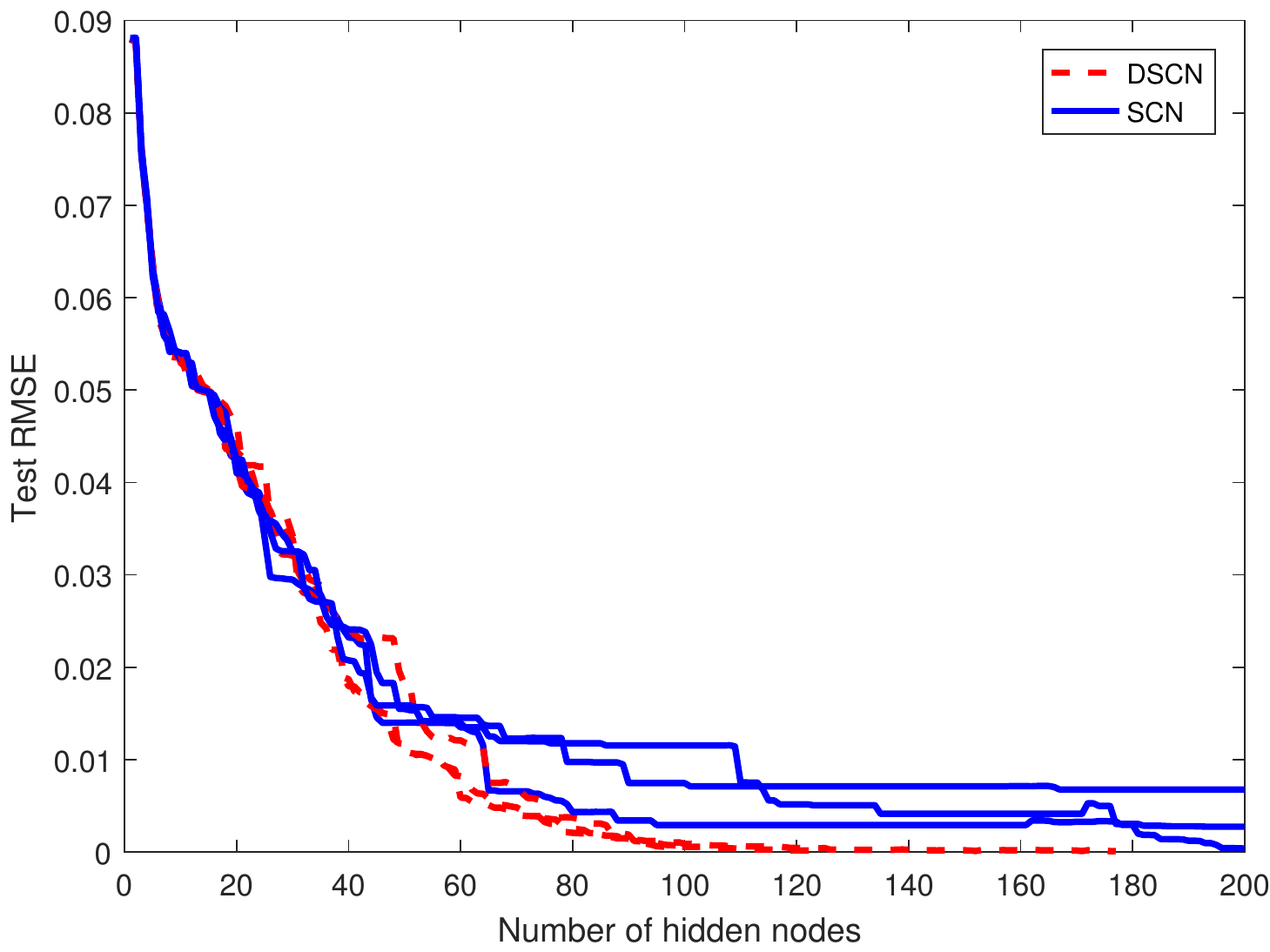}}
\caption{Effectiveness and consistency}
\end{figure}

\subsection{Algebraic Property of the Hidden Output Matrix}
In a common sense, a learner model's capacity can be measured by using a trade-off metric over the learning and generalization performance.  In addition, we may evaluate the capacity of a learner model by looking at the interpretability. In fact, it is always desirable to have a learner model with balanced performance among learning, generalization and model complexity. For  DeepSCN framework, due to the specific configuration in model building, the model capacity is not only associated with the number of nodes for each hidden layer, but also closely related to some algebraic properties  of the hidden output matrix.  Technically, we found that the degree of the rank-deficiency of the joint hidden output matrix is proportional to the level of redundancy.  To do so, the degree of rank-deficiency is computed by $p=\mbox{rank}(\mathcal{H})/\sum_{n=1}^jL_n$, where $\mathcal{H}$ represents the hidden output matrix composed of all existing hidden nodes, $\sum_{n=1}^jL_n$ represents the (current) number of hidden nodes after adding $j (j=1,2,3,4)$ hidden layers and $L_j$ hidden nodes $1\leq L_j\leq 25$. The values of $p$ for both SCN and DeepSCN are plotted in Figure 3(a), in which it can be observed that $\mathcal{H}$ of DeepSCN is with higher rank even when the number of hidden nodes is approaching to 100, however, the degree of the rank-deficiency for the SCN becomes much more higher as $L$ becomes larger. Note that the results from DeepSCN reported in Figure 3(a) were obtained with a fixed architecture, i.e., 4 hidden layers with 25 nodes for each. Figure 3(b) depicts six curves of the rank-deficiency for both the SCNs and DeepSCNs, where $M=4$ and  $L_{max}^{(n)}=50$ for each hidden layer are used in the DSCN algorithm. From these observations, we can infer that the DeepSCN models  have much less redundant basis functions compared against the SCN models.
\begin{figure}[htbp]
\centering
\subfigure[Rank-deficiency property]{\includegraphics[width=0.46\textwidth]{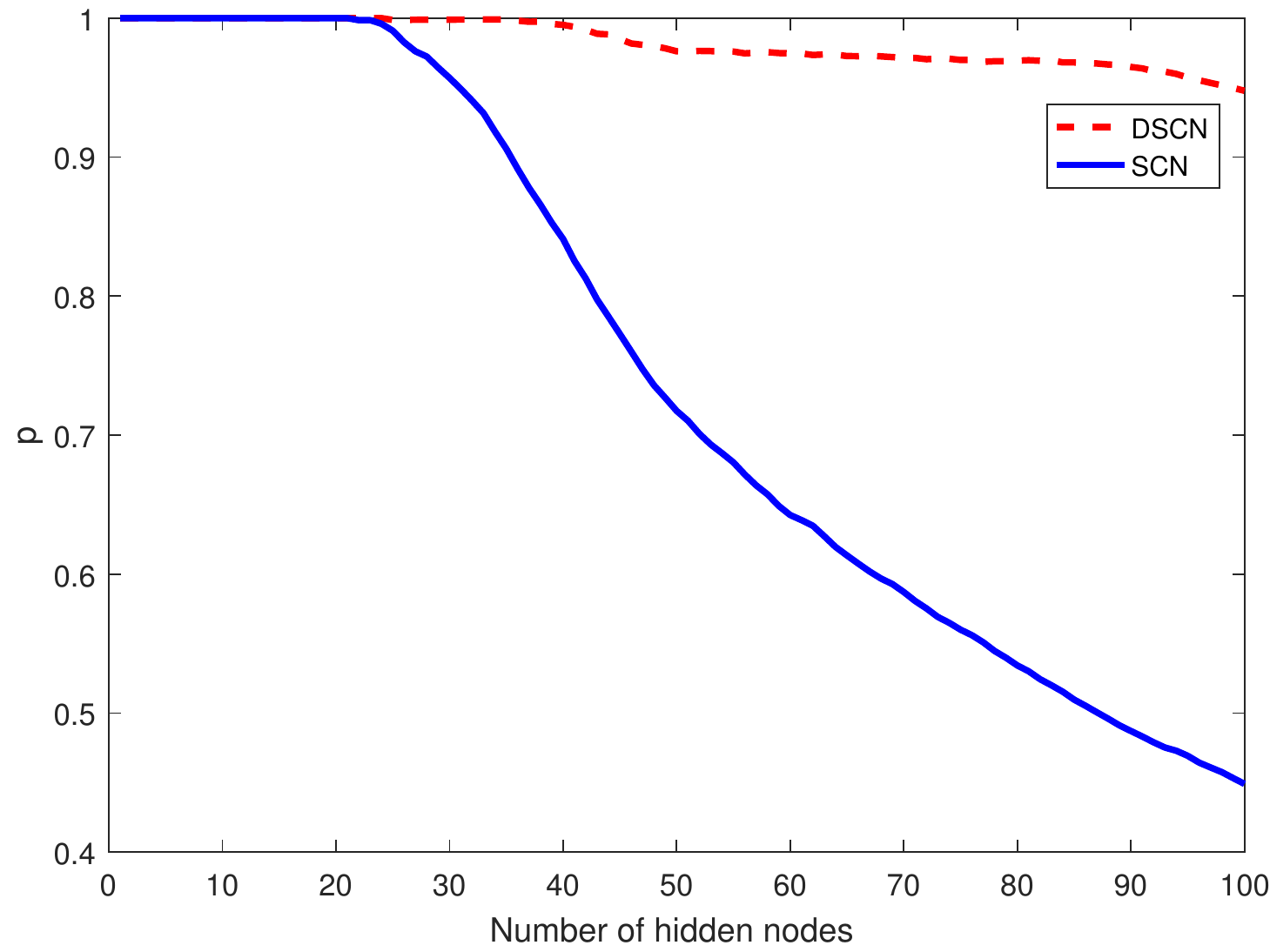}}
\subfigure[Rank-deficiency property]{\includegraphics[width=0.46\textwidth]{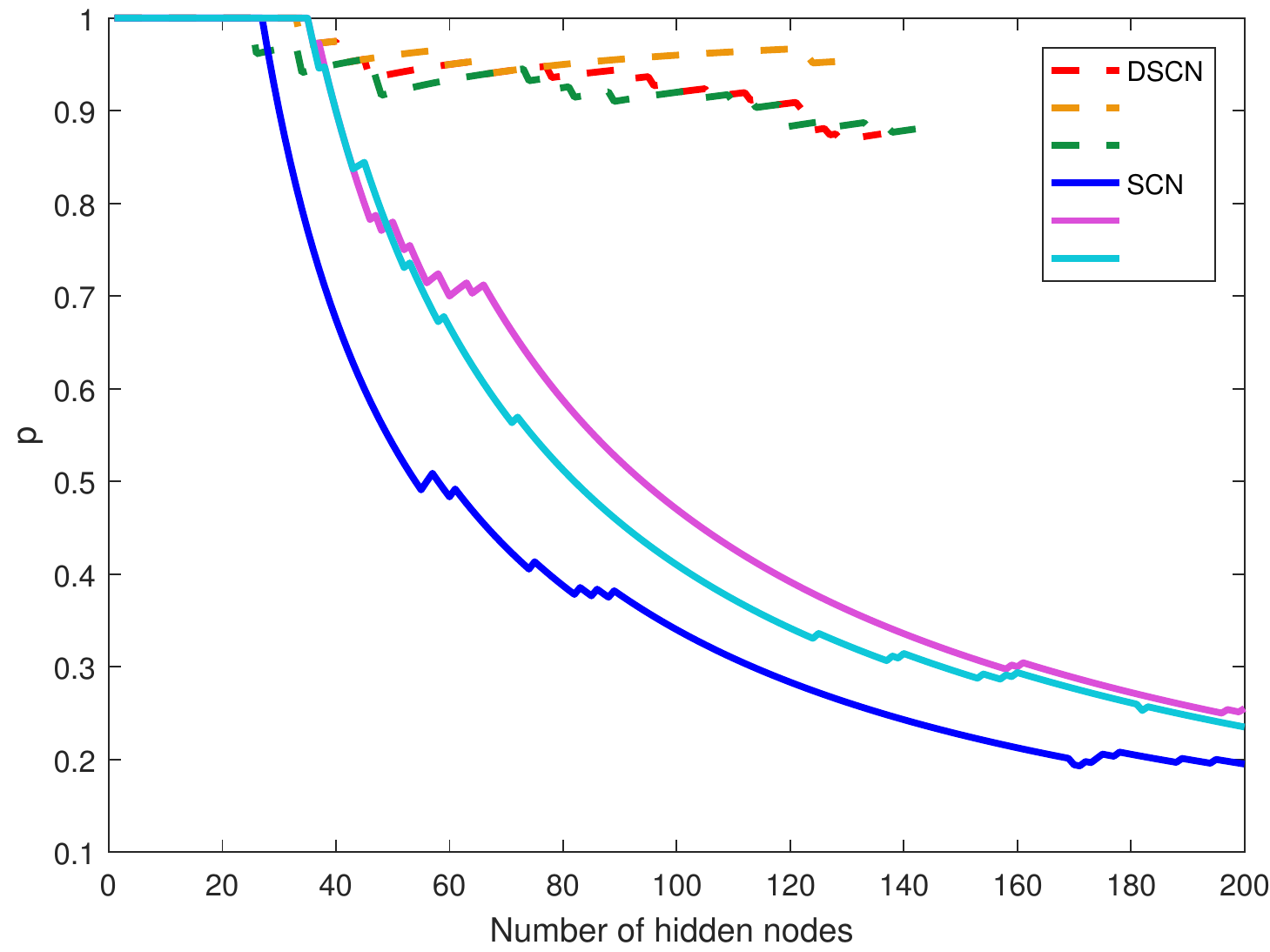}}
\caption{(a) SCN with $L=1:100$ and DeepSCN with $M=4, 1\leq L_j\leq 25$, $j=1,2,3,4$; (b) SCN with $L=1:200$ and DeepSCN with $M=4, 1\leq L_j\leq 50$, $j=1,2,3,4$}
\end{figure}

\subsection{Robustness Analysis}
In general, robustness refers to  the ability of a system to maintain its performance whilst subjected to noise in external inputs, changes to internal structure and/or shifts in parameter setting. Obviously, the quality or performance of DeepSCNs depends upon the parameter setting.  In order to investigate  the robustness of DSCN algorithm, in this paper we limit our study on the learning parameter  $r$  with 11 values only. A similar robustness analysis on another set of key learning parameters $\Upsilon\!=\{\lambda_{min}\!:\!\Delta\lambda\!:\!\lambda_{max}\}$ is not reported here. Figure 4 shows the training and test performances for both SCN and DeepSCN with three different settings of $r$, that is, randomly taking  10 real numbers from the open interval $(0.9, 0.99)$ and arranging them in increasing order to form a set $r_1$, and set $r=\{r_1,1-10^{-6}\}$. As  can be seen that   both SCN and DeepSCN perform robustly against this learning parameter setting (i.e., not sensitive to the uncertain sequence $r$). In this case, we fixed the architecture of DeepSCN with 4 hidden layers and 25 nodes for each. A remarkable drop-down can be observed after completing the construction of the first layer, which clearly exhibits the advantage of DeepSCN over SCN.
\begin{figure}[htbp]
\centering
\subfigure[Training]{\includegraphics[width=0.46\textwidth]{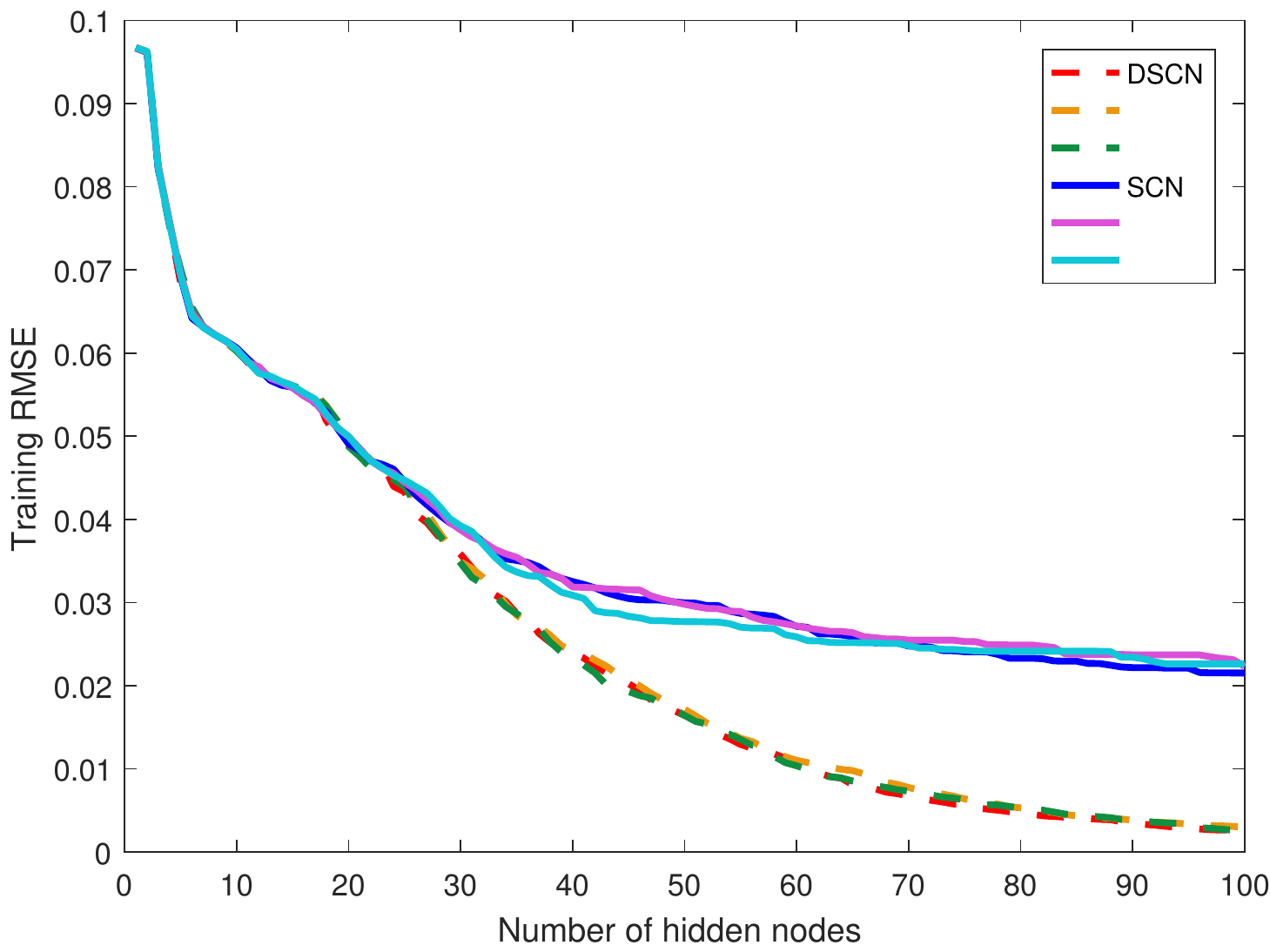}}
\subfigure[Test]{\includegraphics[width=0.46\textwidth]{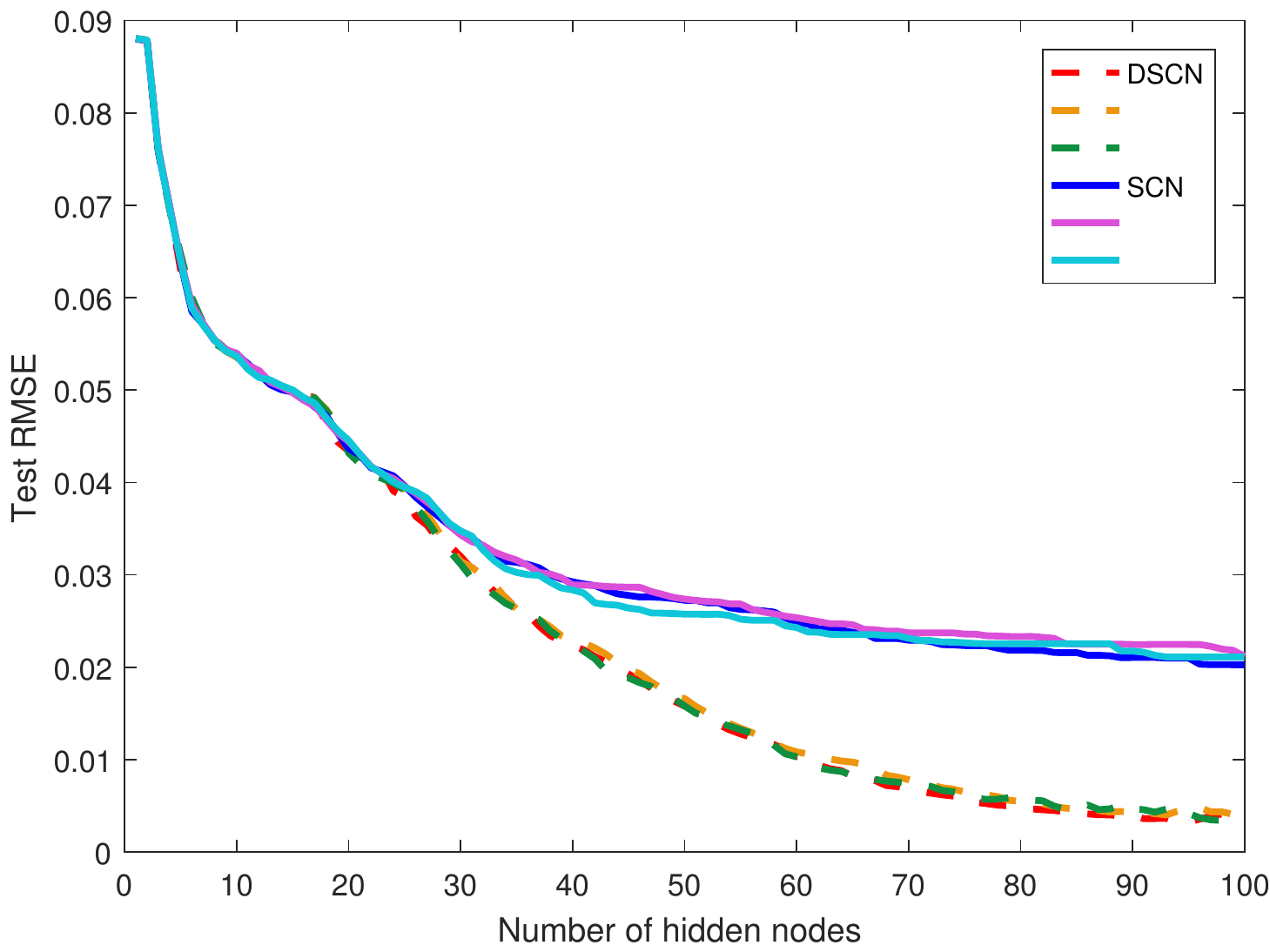}}
\caption{Robustness analysis on the learning parameter setting}
\end{figure}

\section{A Case Study: Rotation Angles Predication for Handwritten Digits}
This section further investigates the merits of the proposed DeepSCN by  predicting the angles of rotation of handwritten digits. We have a collection of synthetic handwritten digits containing 5000 training and 5000 test images of digits with corresponding angles of rotation. Each image represents a rotated digit in grayscale and of normalized size ($28 \times28$). Figure 5 displays 16 random sample training digits, retrieved from Neural Network Toolbox in MATLAB R2017b.
\begin{figure}[htbp]
\centering
\includegraphics[width=0.4\textwidth]{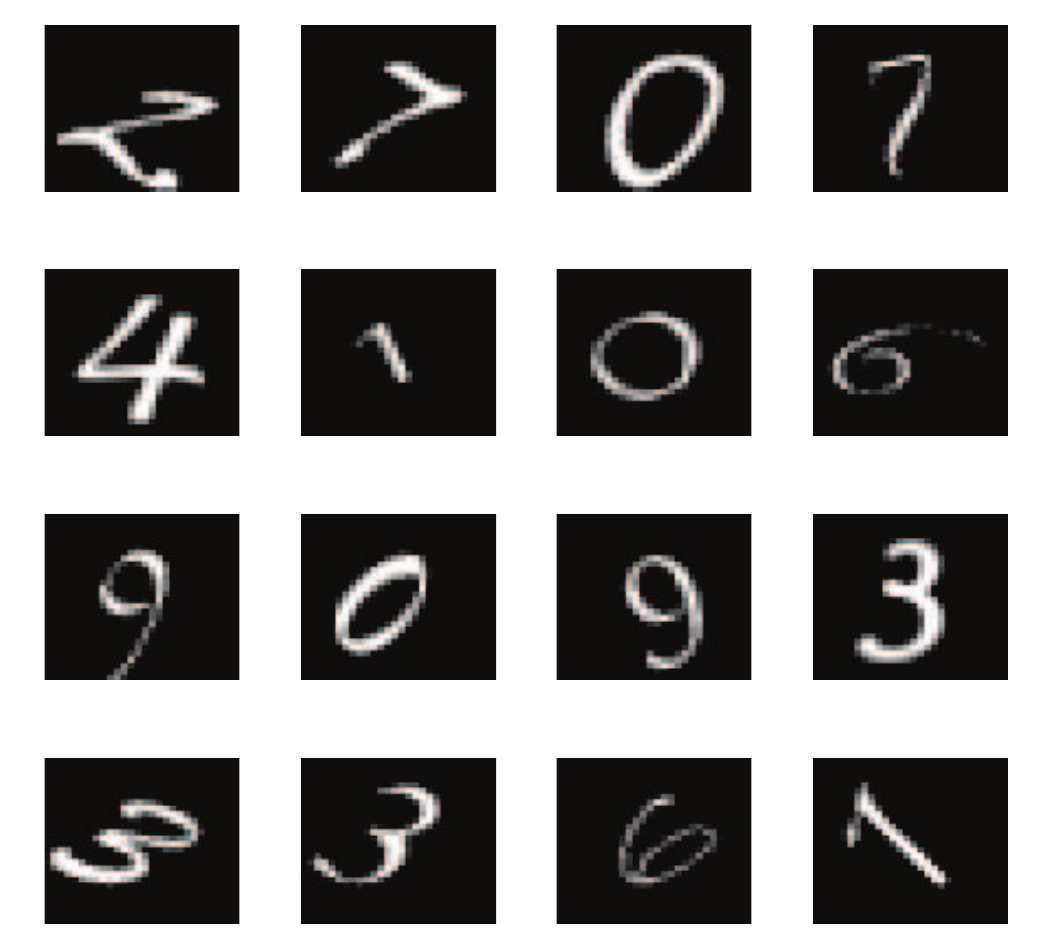}
\caption{Samples of Handwritten Digits with Angles of Rotation}
\end{figure}

To demonstrate the advantages of DeepSCN over SCN framework, we employ two kinds of evaluation metric in the performance comparison: (i) The Percentage of Predictions within an Acceptable error margin (PPA); and (ii) The root-mean-square error (RMSE) of the predicted and actual angles of rotation.  In particular for measuring PPA, we need to calculate the prediction error between the predicted and actual angles of rotation, followed by counting the number of predictions within an acceptable error margin from the true angles. In our experiments, we set the threshold to be 10 degrees. Therefore, the PPA value within this threshold can be obtained by
\begin{equation*}
  PPA=\frac{\#\{|Prediction\:\:Error| < 10\}}{Number\:\:of\:\:Sample\:\:Images}.
\end{equation*}

\begin{table}[htbp!]
\renewcommand{\arraystretch}{1.35}
	\caption{Performance Comparison}
	\label{calssification_table}
	\centering
	\begin{tabular}{|c|c|c|c|c|}
		\hline
		\multirow{2}{*}{Algorithms} & \multicolumn{2}{c|}{PPA ($\%$)} & \multicolumn{2}{c|}{RMSE} \\
		\cline{2-5}
		& Training & Test & Training & Test \\
		\hline
		$\mbox{DSCN}^{1}$ &	\textbf{82.21} & \textbf{65.21}  &\textbf{7.0265} & \textbf{14.8231}   \\
		\hline
		$\mbox{SCN}^{1}$  &78.14 &60.35 &7.5623 &  15.9176 \\
		\hline
$\mbox{DSCN}^{2}$ &\textbf{84.73}	& \textbf{66.36}& \textbf{5.2324}& \textbf{12.0971}   \\
		\hline
		$\mbox{SCN}^{2}$  &80.12  & 60.00 & 5.6797 & 14.7775  \\
		\hline
\multicolumn{5}{|l|}{\scriptsize \tabincell{c}{$L=1000$ for $\mbox{SCN}^{1}$, $L_1\!=L_2\!=L_3\!=L_4\!=250$ for $\mbox{DSCN}^{1}$\\
$L=2000$ for $\mbox{SCN}^{2}$, $L_1\!=L_2\!=L_3\!=L_4\!=500$ for $\mbox{DSCN}^{2}$}}\\
\hline
	\end{tabular}
\end{table}
It is logical to think that higher PPA and/or lower RMSE values can reflect a better capability on this regression task. In Table I, we summarize the training and test results for both DSCN and SCN to illustrate the superiority of our developed DeepSCN framework. Specifically, two options for the architecture setting are examined, that is, $L=1000$ for SCN, $L_1=L_2=L_3=L_4=250$ for DSCN, and $L=2000$ for SCN, $L_1=L_2=L_3=L_4=500$ for DSCN, respectively. As can be seen in Table I, DSCN outperforms SCN in these two cases with higher PPA and lower RMSE values. It should be noted that all the results shown here are the averaged values based on five independent trials. Based on our experience, their standard deviations have no significant difference, so we omit this information here without any confusion.
\begin{figure}[htbp]
\centering
\subfigure[SCN: $L=2000$]{\includegraphics[width=0.46\textwidth]{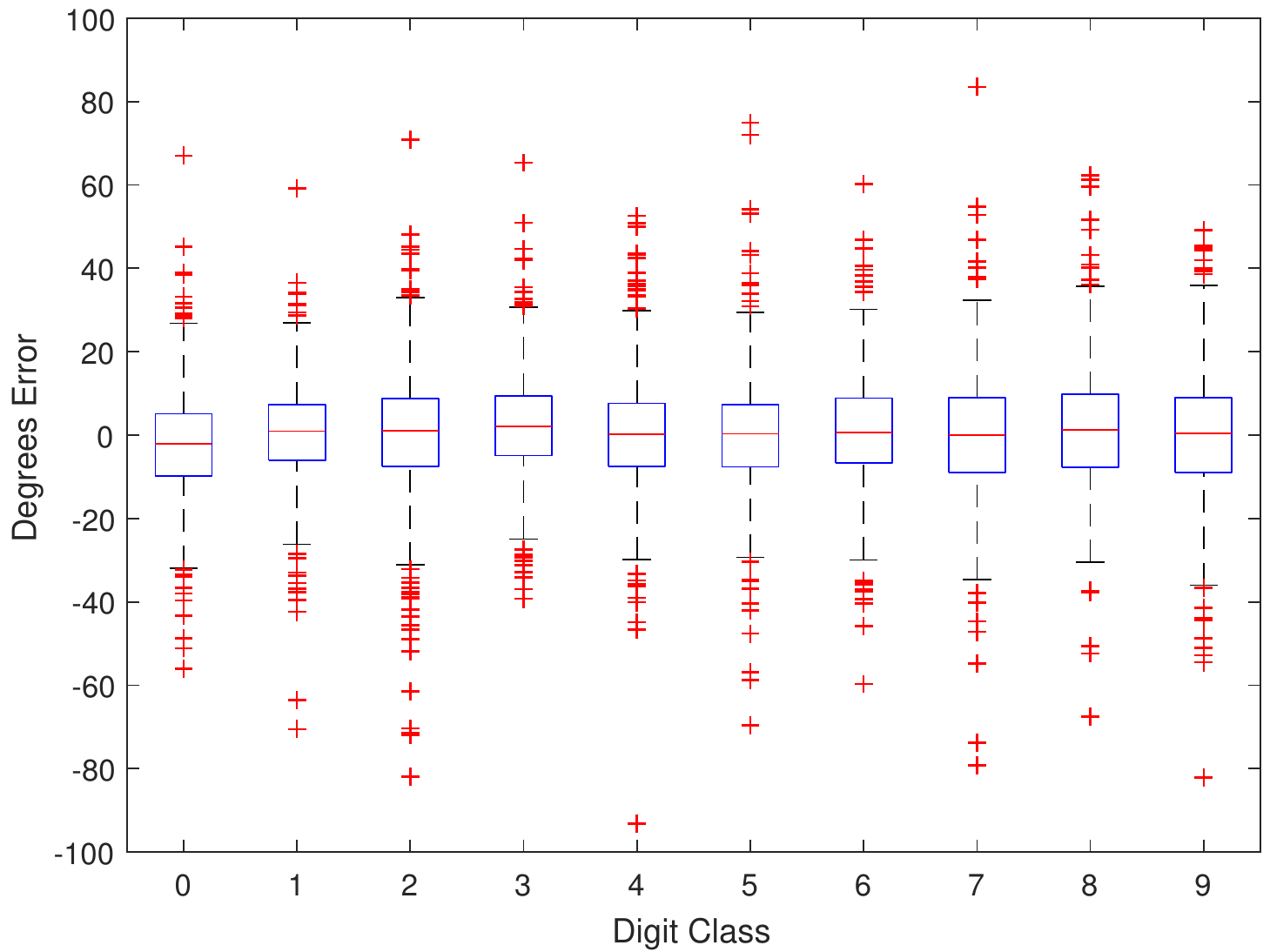}}
\subfigure[DSCN: $L_1=L_2=L_3=L_4=500$]{\includegraphics[width=0.46\textwidth]{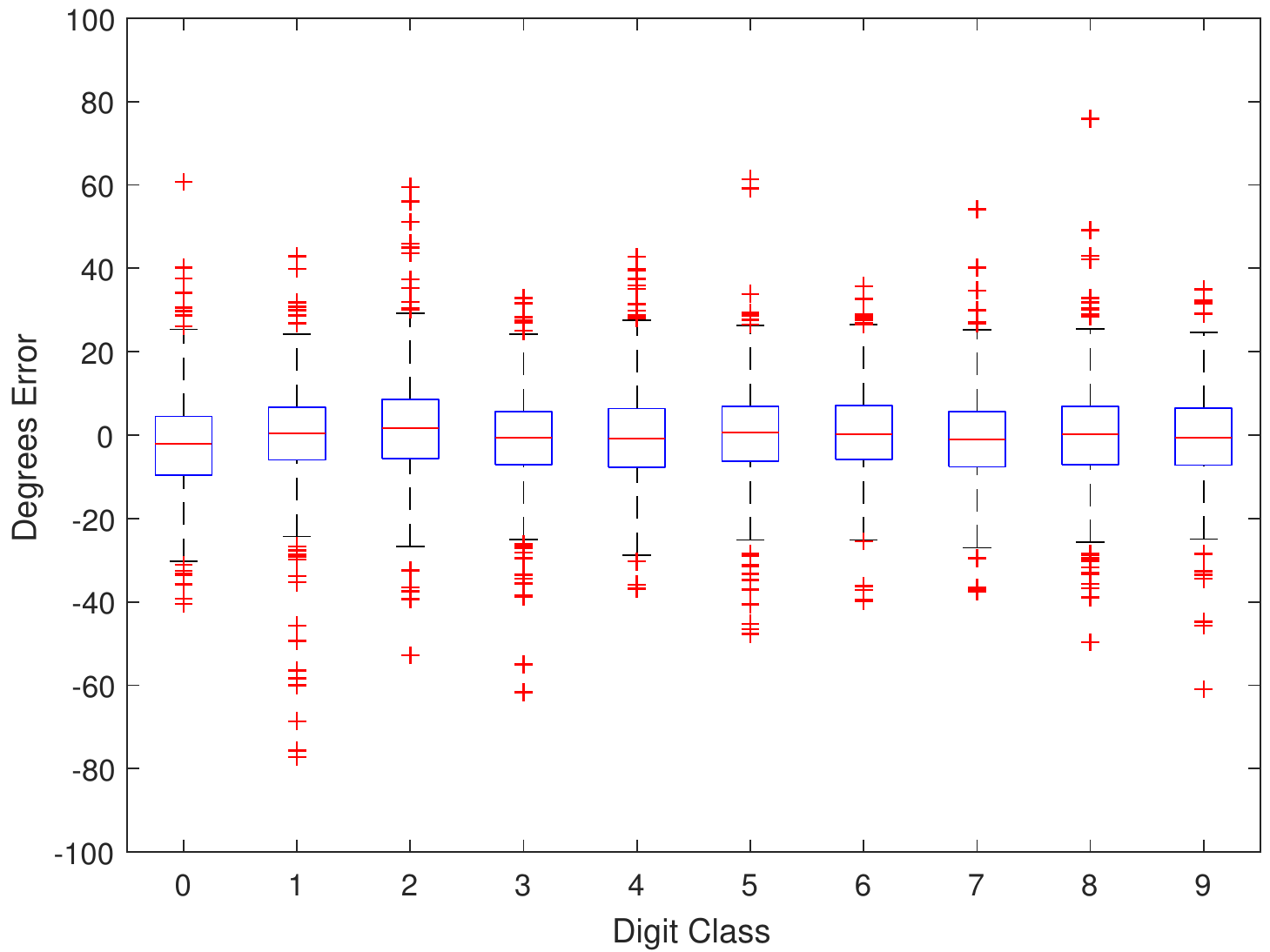}}
\caption{Display Box Plot of Residuals for Each Digit Class}
\end{figure}

In Figure 6, we use the box-and-whisker diagram to show the difference between DSCN and SCN on residual distribution for test images (for each digit class). Intuitively and obviously, DSCN works much more favorably than SCN as there are less records with abnormal degree error, as marked in red plus symbol. In particular, for the digit '1', '2', '5', '7', '8', DSCN shows more robustness in predicting the associated angles of rotation. The digit classes with highest accuracy have a mean close to zero and little variance, such as '3' and '6'. The learner model produced by SCN is more prone to result in larger predicted angles, i.e, a plenty of examples with degree error outside the interval [-60,60]. With consideration of length limitation, we only provide the results of the architecture setting with $L=2000$ for SCN, $L_1=L_2=L_3=L_4=500$ for DSCN. Based on our simulations, similar findings can be found in the case that $L=1000$ for SCN, $L_1=L_2=L_3=L_4=250$ for DSCN.
\begin{figure*}[htbp]
\centering
\subfigure[Original]{\includegraphics[width=0.31\textwidth]{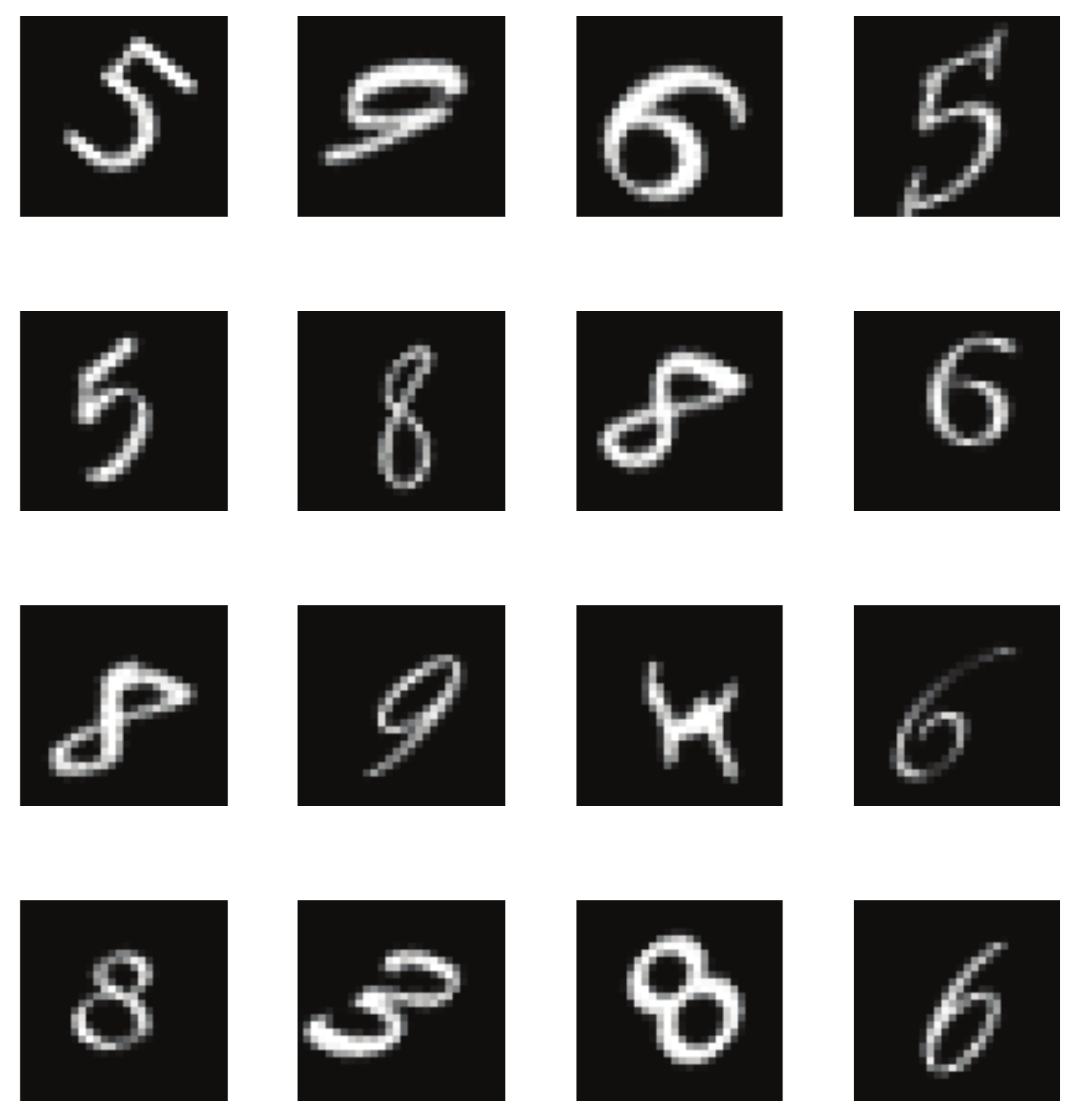}}\hspace{4mm}
\subfigure[Corrected by SCN]{\includegraphics[width=0.31\textwidth]{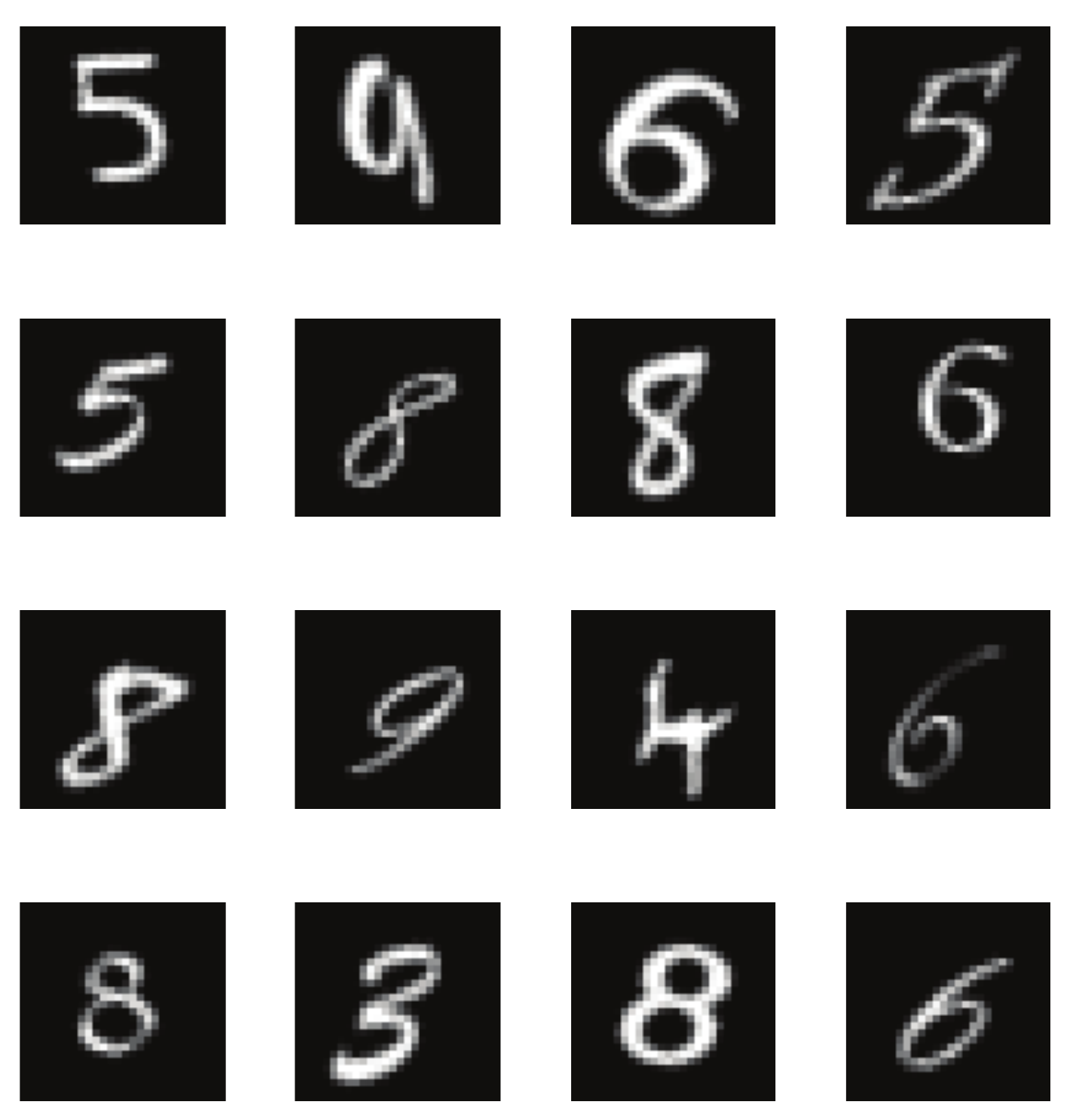}}\hspace{4mm}
\subfigure[Corrected by DSCN]{\includegraphics[width=0.31\textwidth]{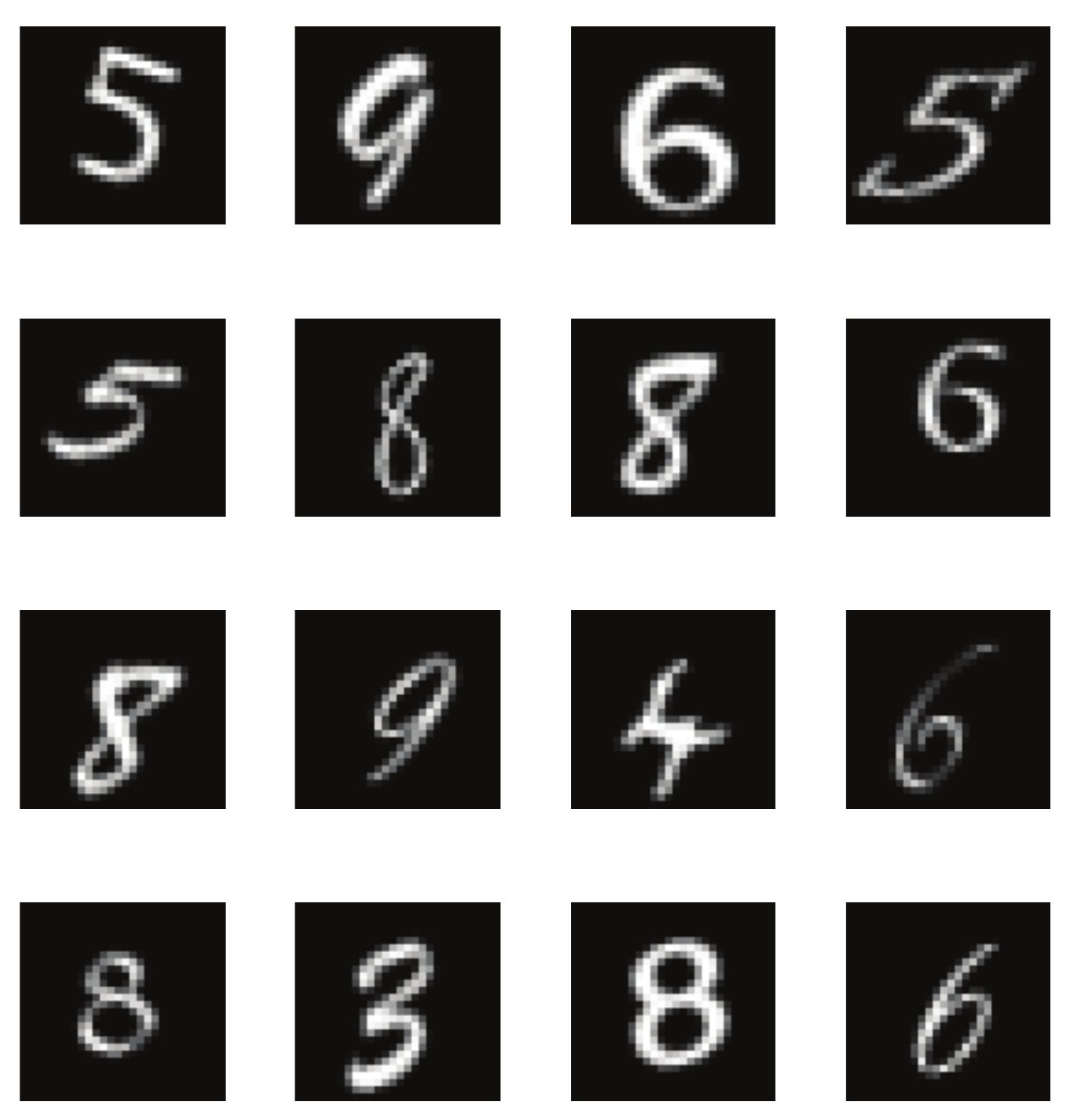}}
\caption{Demonstration of Original Digits and Their Corrected Rotations: (b) $L=2000$ for SCN; (c) $L_1=L_2=L_3=L_4=500$ for DSCN.}
\end{figure*}

With the predicted results, we can rotate the sample digits according to their predicted angles of rotation, aiming to correct the digits into a regular position. In Figure 7, we randomly selected 16 test samples and display their original images, the corrected digits by SCN, and the corrected ones by DSCN, respectively. It is obvious that DSCN can contribute to a better  rotation correction than SCN, showing that the proposed DeepSCN framework have more advantages over our previously developed SCNs \cite{wang}. So far, we have extensively compared DSCN with SCN on this specified data modelling task. As for the paper length limitation, we have no extra space to add more robustness analysis on these two algorithms. Additionally, more in-depth theoretical analysis (in terms of our innovative SCN methodology) why deep is better more shallow are desirable in our future work.

\section{Concluding Remarks}
Although many empirical proofs demonstrate great potential of deep neural networks for learning representation, it is still blind for  end-users to deploy the network architecture so that the resulting deep learner model has sufficient capacity to approximately represent signals in some forms. Except for concerns on the architecture determination, one cares much about  learning performance in terms of effectiveness and efficiency.  In this paper, our proposed DeepSCNs, as a class of randomized deep neural networks, offer a fast and feasible pathway for problem solving.  Our technical contributions to the working field can be summarized as follows:
\begin{itemize}
\item The proposed supervisory mechanism (\ref{step2})  is the key for developing DeepSCN framework, which could be used as a unique feature to distinguish our work from other randomized learner models;
\item The scopes of random parameters at each layer of DeepSCNs could be updated adaptively, which make the set of basis functions  rich and functional;
\item  The construction of  DeepSCNs is data dependant, and the proposed  algorithm has much less human intervention provided that a set of appropriate learning parameters is given;
\item The removal of the redundant nodes becomes possible with some post-processing steps, which implies that we can make a better learning representation with sparsity.
\end{itemize}

There exist a lot of opportunities to further develop the proposed framework in either theoretical aspects or real-world applications. For instance, one can explore some algebraic properties of the DeepSCN to evaluate the quality of the resulting model,  seek for different forms of  supervisory mechanisms, develop some advanced learning schemes for random representation, study convolutional or recurrent  versions of the proposed DeepSCN. For applications,  the DeepSCN model has great potential for extracting random features with lower computational cost and fast implementation, which is very important for big data classification with hierarchical or cascade structure.

Before ending up this paper, we would like to share our philosophy behind developing randomized methods for DNNs. The territory of the weights and biases of the randomized learner models must be data-dependant and it can be characterised by certain constraints in relation to the model's capacity. From our understandings on the randomized learner's quality, it is highly desired to make the joint hidden output matrix full-rank. Unfortunately, such an essential algebraic condition  may not be achieved  by simply assigning the random hidden parameters from  a fixed scope such as [-1,1] or a pre-defined probability distribution. Therefore,  the way to generate the random parameters at the hidden layers with guarantee for both universal approximation property and sound generalization capability  should be done.


\end{document}